# Old wine in old glasses

# Comparing computational and qualitative methods in identifying incivility on Persian Twitter during the #MahsaAmini movement

Hossein Kermani
Political communication group, Department of communication, University of Vienna
Hossein.kermani@univie.ac.at

Fatemeh Oudlajani, Allameh Tabataba'i University, fatemehoudlajani@gmail.com

Pardis Yarahmadi, Queensland University of Technology, P.Yarahmadi@hdr.qut.edu.au

Hamideh Mahdi Soltani, Allameh Tabataba'i University, ms.soltani1991@gmail.com

Mohammad Makki, University of Wollongong, Australia, mmakki@uow.edu.au

Zahra HosseiniKhoo, University of Vienna, z.hosseinikhoo@gmail.com

Abstract

This paper compares three approaches to detecting incivility in Persian tweets: human qualitative coding, supervised learning (ParsBERT), and large language models (ChatGPT). Using 47,278 tweets from the #MahsaAmini movement in Iran, it evaluates the accuracy and efficiency of each method. ParsBERT significantly outperformed all seven studied ChatGPT models in identifying hate speech. The results also show that ChatGPT struggles not only with subtle messages but also with detecting incivility in more explicit content. Results also show that using a prompt with different languages (English vs. Farsi) has no remarkable effect on ChatGPT's output. This study offers a detailed comparison of these methods, contributing to our understanding of their effectiveness and nuances in analyzing hate speech in a low-resource language.

'The path of love seemed easy at first, but what came was many hardships.' This is probably the most famous poem by Hafiz, a great Iranian poet, whose book of poems, Divan, starts with. This hemistich probably describes using computational methods in communication studies as well. The introduction of automated techniques raised much optimism about their potential to transform communication research. All this enthusiasm was not in vain. Computational methods helped researchers to analyze big data in less time and with fewer human and financial resources (Barbera & Steinert-Threlkeld, 2020). Nevertheless, after a short time of optimism about 'big data,' which 'speaks for itself,' (Anderson, 2008) researchers came to understand the serious challenges of using computational techniques in social research (boyd & Crawford, 2012; De Grove et al., 2020). For instance, scholars have raised critical concerns about the validity and reliability of automated techniques (Nicholls & Culpepper, 2020). The usefulness of computational methods in communication studies is no longer a question. The exponential growth of data and the popularity of online services make using them inevitable. The question now is when and under which circumstances these methods could provide reliable and valid results. This paper aims to contribute to this line of inquiry by comparing human-driven qualitative analysis with different computational methods (BERT and ChatGPT models, i.e., ChatGPTs) in the identification of a pertinent problem in the social media environment: incivility, i.e., hate speech.

Social media was once thought of as spaces where healthy communication could be forged and muted groups could raise their voices (Tufekci, 2017). However, incivility is a threat to the democratic potential of social media. In recent years, the presence of online hate speech has increased despite the various control measures in place to restrict it. Surveys across several countries indicate that 42%–67% of young adults observed 'hateful and degrading writings or speech online' (Keipi et al., 2016; Räsänen et al., 2016) and 21% have been victims themselves (Oksanen et al., 2014). Incivility could diminish political trust and result in polarization (Coe et al., 2014). In addition, online hate has negative effects on the well-being of both victims and observers (Walther, 2022). Computational methods are considered effective ways to identify and combat online hate and incivility on a large scale. Researchers used different automated techniques to detect hate speech on social media. In particular, NLP (Natural Language Processing) tasks have become popular in this attempt. Dictionary-based techniques, Supervised Machine Learning (SML) algorithms like BERT, and recently introduced Large Language models (LLMs) like ChatGPT were employed in automated hate speech detection (Kovács et al., 2021; Saleh et al., 2023a).

Despite this remarkable line of inquiry, there are still some gaps in our understanding of automated hate speech detection. First, most studies focus empirically on mainstream languages like English and German (Kovács et al., 2021; MacAvaney et al., 2019a; Stoll et al., 2020). There is a severe lack of research in other languages like Farsi. Studying other languages not only contributes to our knowledge in different contexts; but it could also be a step forward in multi-lingual NLP, particularly automated hate detection (Lind et al., 2022). In addition, this line of inquiry mainly analyzes offensive vs. non-offensive messages (Pendzel et al., 2023; Struppek et al., 2024). This binary approach neglects the other types of uncivil tweets, in addition to overlooking more subtle forms of hate. Identification of subtle forms of incivility remains a challenge to date (Davidson et al., 2017; Stoll et al., 2020). This problem is intensified when it comes to margin languages like Farsi and rhetorical forms of incivility. We will analyze the strengths and weaknesses of computational methods in the automated detection of implicit incivility. Moreover, the existing literature examined different Deep Learning (DL) and Machine Learning (ML) methods, but the use of recently introduced Large Language Models (LLMs) is new (Albladi et al., 2025). Most studies employed early versions of ChatGPT or other LLMs like GPT-3.5 Turbo (Albladi et al., 2025; Delbari et al., 2024). In particular, there are not many studies on HS detection in Persian languages using LLMs. We will address this fallacy by using different advanced ChatGPT models, like o1 and 4.1, and comparing the results with discursive interpretations and BERT as a supervised model. We will also go further to investigate the effect of the prompt's language (Farsi vs. English) on ChatGPTs' output.

This paper is structured as follows. First, we discuss the theoretical aspects of incivility, followed by a review of the existing research into automated HS detection, particularly in Farsi. In this section, we highlight the gaps in the research that the paper tries to address. Next, we discuss the context, data, and methods of the study. Findings will follow, and finally, we will discuss the contribution of this research. The article mainly enhances our understanding of automated HS detection by revealing that old SML models are still more powerful than ChatGPTs. However, the performance of ChatGPTs in the detection of implicity is better than the detection of incivility. We also argue that using prompts in different languages does not affect the results significantly.

**Theoretical debates: different faces of incivility**

From a general standpoint, incivility is defined as impolite behavior that violates the polite norms of interpersonal communication (Coe et al., 2014; Gervais, 2014; Rains et al., 2017). Hameleers et al. (2022) argue that incivility is any type of offensive statement that trespasses the ideal type of democratic communication. Coe et al.'s (2014) definition of incivility has been cited in many works. They understand incivility as features of discussion that convey an unnecessarily disrespectful tone toward the discussion forum, its participants, or its topics (p. 660). Hate speech (HS) is another buzzword in this area (Hietanen & Eddebo, 2023). Hate speech could be understood as a radical form of incivility (Hameleers et al., 2022); however, hate speech and incivility have been used interchangeably in many studies (Hietanen & Eddebo, 2023; MacAvaney et al., 2019b; Saleh et al., 2023b). We will follow this line of study in this paper. More specifically, the European Commission (2019) defined HS as a language that incites violence or hatred against a group, defined in relation to race, religion, or ethnicity. Since there are many overlaps and mutual conceptualizations in these definitions, it is plausible to consider HS and incivility interchangeable for the sake of this research.

There are different classifications of incivility as well. Coe et al. (2014) suggest a typology of uncivil speech: Name-calling (mean-spirited or disparaging words directed at a person or group of people), Aspersion (mean-spirited or disparaging words directed at an idea, plan, policy, or behavior), Vulgarity (using profanity or language that would not be considered proper in professional discourse), Lying (stating or implying that an idea, plan, or policy was disingenuous), and the pejorative for speech (disparaging remark about the way in which a person communicates) (p. 661). These scholars used this typology in coding newspaper website comments. While this typology is widely accepted, there are some reservations about using it to code social media data, particularly in NLP tasks. First, name-calling, vulgarity, and aspersion are highly overlapping. In many tweets in our dataset, we found instances using profane language to attack individuals, groups, or ideas. This fact will probably affect the performance of machine models and cause them not to differentiate between these classes precisely. Also, other scholars did not mention lying as a type of incivility or hate speech. Lying, particularly after the 2016 U.S. presidential election, is studied as a form of dis/misinformation (Freelon & Wells, 2020). Finally, this classification does not consider the levels of incivility. There are tweets that are more uncivil as they may embed higher levels of profane language, while others may not.

The Anti-Defamation League (ADL) proposes five levels in the form of a pyramid, distributed from the lowest to the highest intensity: stereotypes, insults, discriminatory expressions, and genocide (Paz-Rebollo & Segado.Boj, Forthcoming). This classification solves the problem of intensity. Still, the challenge of insulting language vs. non-insulting but uncivil speech persists. In other words, some uncivil messages may include no insulting words. Such rhetorical forms of incivility are significant in the computational identification of hate speech. Thus, the significance of discriminating between them. Davidson et al.'s (2017) differentiation between hate speech and offensive language could be a solution. They define hate speech as language that is used to express hatred towards a targeted group or is intended to be derogatory, humiliating, or insulting to the members of the group. Then, they include offensive tweets in their analysis as messages include clear bad words. Drawing on this line of research, we classify incivility into three categories: Pejorative Speech (PS), Insult, and Threatening Messages (TMs). Unlike Coe et al.'s (2014) definition of pejorative for speech, our definition of PS targeted words or grammatical forms expressing a negative or disrespectful connotation, a low opinion, or a lack of respect toward someone or something. It is also used to express criticism, hostility, or disregard. PS could include stereotypes or discriminatory language against groups or individuals. While PS could employ or not employ rhetorical devices, it does not include profane language. Therefore, it can be said that there is probably a clear distinction between it and aspersion and name-calling in Coe et al.'s (2014) typology. However, PS is sharply distinguished from vulgarity. Vulgar messages could be understood as an insult, which echoes the ADL concept. This kind of hate speech specifically uses bad words to ridicule and denigrate its targets. Finally, we discriminate between insults and threats, as the latter have received particular attention in research (Hietanen & Eddebo, 2023; Paz-Rebollo & Segado.Boj, Forthcoming). Threatening messages are understood as an extreme form of hate speech. We hypothesize implicity as an additional dimension of all three types. Therefore, each of these types could be implicit or explicit. For instance, Insult could be direct (*fuck, whore*) or indirect and more rhetorical (*no one knows who your father is*). We pay particular attention to the capability of computational methods in identifying implicit hateful messages, as there is still a gap in research (Albladi et al., 2025; Stoll et al., 2020).

**Computational identification of incivility: From small bags to large transformers**

Researchers employed a wide range of computational methods to identify incivility. Dictionary-based models have been implemented in early research to detect uncivil messages (Davidson et al., 2017). However, due to the fundamental issues of Bag-of-Words (BoW) models, such as their inability to understand more complex forms of incivility, they soon become outdated (Stoll et al., 2020; Van Atteveldt et al., 2022). Researchers employed other techniques, e.g., deep neural networks such as Long Short-Term Memory (LSTM) and Convolutional Neural Network (CNN), or supervised classification algorithms such as Support Vector Machines (SVM) independently or in combination with BoW and other algorithms to improve the output (Badjatiya et al., 2017; Gambäck & Sikdar, 2017; Malmasi & and Zampieri, 2018; Zhang et al., 2018). DL algorithms, however, suffer from some weaknesses, particularly issues like vanishing gradients and difficulties in managing long-term dependencies (Eren & Küçükdemiral, 2024).

Nonetheless, transformer-based language models open a whole different window to many NLP tasks like text classification. Bidirectional Encoder Representations from Transformers (BERT) (Devlin et al. 2018) is a language model trained on huge data based on contextual representations. BERT introduced the concept of bidirectional training, where models are trained to consider both the left and right context in all layers, which led to improving performance on different NLP tasks like hate speech detection (Pàmies et al., 2020). For instance, Zhu et al. (2019) fine-tuned the BERT model for this task and achieved an 83:88% F1-score in classifying hateful tweets. In combination with a variety of classifiers, including CNN, Mozafari et al. (2020) found that BERT provided the highest F1-score, which is 92%. Other fine-tuned BERT models like ALBERT (Wiedemann et al., 2020), RoBERTa (Safayani et al., 2024), mBERT (Pérez et al., 2020), and XLM-RoBERTa (Wang et al., 2020) have also been employed for this aim.

LLMs are new players in the field that have attracted my attention in recent years. The boom of LLMs like ChatGPT has heavily affected the field of automated text classification. LLMs are capable of successfully performing many language-processing tasks zero-shot (without the need for training data) or few-shot. Early studies like Gilardi et al. (2023) claim that ChatGPT outperforms MTurkers on a variety of annotation tasks. Reiss (2023), on the other hand, finds that LLMs perform poorly on text annotation and argues that this tool should be used cautiously. Numerous other studies present similar analyses with varying results and recommendations (e.g., He et al., 2023; Wang et al., 2021; Ziems et al., 2023; Zhu et al., 2023; Ding et al., 2022). These studies have been mainly conducted on popular, publicly available benchmark datasets, comparing the LLMs' results with the work of crowdsource annotators. First, these tests are plausibly affected by contamination (Brown et al., 2020), meaning that the datasets may be included in the LLM's training data and that strong performance may reflect memorization, which will not generalize to new datasets and tasks. Second, as discussed above, crowdsource annotators often suffer from a limited attention span, fatigue, and changing perceptions of the underlying conceptual categories throughout the labeling procedures (Grimmer and Stewart, 2013; Neuendorf, 2016). Ethical concerns in employing LLMs have been raised as well (O'Neill & Connor, 2023).

In addition, LLMs in hate speech detection are new, and we are far from a substantial line of research in this field. González-Bustamante (2024) tested different models, including GPT-4o, Nous Hermes 2 Mixtral, and LLaMA, and revealed that GPTs tend to show not only excellent computing time but also overall good levels of reliability. In another study, Roy et al. (2023), testing different models on three large datasets, find that, on average, including the target information about hateful language in the pipeline improves the model performance substantially (∼ 20−30%) over the baseline across the datasets. Their study was a step forward from previous research (e.g., Huang et al., 2023) in which their prompt was framed as a 'yes/no' question (rather than based on the exact classes). Other researches also highlight the power of LLMs, particularly ChatGPTs, in the automated identification of incivility (Deroy & Maity, 2023; Leidinger et al., 2023). Nonetheless, this line of inquiry suffers from several gaps, like its limitation to a binary classification of hateful messages (hateful vs. non-hateful) or restriction to early versions of ChatGPT models (Pendzel et al., 2023; Struppek et al., 2024). As a result, less attention has been paid to more subtle and rhetorical forms of incivility (Albladi et al., 2025). We should also add that most of the studies in this area, particularly those that compare BERT with ChatGPTs, are published in pre-print platforms like Arxiv or presented in different conferences and have not been peer-reviewed, as evident from the references we used above. Most such studies do not dig deeper into the more complicated layers of uncivil messages and are mainly limited to a quick analysis of the performance of various models in HS detection (Guo et al., 2024; Huang et al., 2023). Furthermore, despite a growing body of literature on other languages, the leading strand of research in this area still focuses on mainstream languages like English. Since the gaps we briefly mention here are valid for studies into the Persian language[1] as well, we discuss them in more detail in the next section.

---

[1] We use Farsi and Persian interchangeably in this paper.

**Automated detection of incivility on Persian data**

There is, fortunately, a growing body of literature on the automated detection of incivility in Persian data. Delbari et al. (2024) provide a public dataset (Phate) of 7,000 Persian tweets and classify them into three classes: Vulgar, Hate, and violence. They reveal that ParsBERT (Farahani et al., 2021) and XML-R worked better (F1-scores 57.8 and 58.3, respectively) in HS detection than other models and even ChatGPT. They employed a GPT-3.5-turbo model, which returned a weak F1-score of 43.5. Using their dataset, an anonymous scholar (2025) tests different LLMs with fine-tuned BERT-base models. They show that ParsBERT is a promising model, particularly when fine-tuned with an enriched dataset resulting from weak labeling. However, they repeated Delbari et al. (2024 )'s findings that GPT 3.5 Turbo does not work well on Persian tweets. Nevertheless, their result indicates that using few-shot models worked better than zero-shot ChatGPT models. Mozafari et al. (2024), in another effort, employed a number of ML, DL, and transformer-based neural networks on a 6,000-tweet dataset to identify offensive messages. They find that ParsBert works well in distinguishing between offensive and non-offensive tweets, while its performance in identifying the targets of hatred messages was weak. Furthermore, Safayani et al. (2024) followed a keyword approach to collect 17,000 Persian tweets and implemented different ML models to identify offensive and non-offensive messages. Their research identifies XLM-RoBERTa as the most efficient model. Finally, Karami Sheykhlan et al. (2023) also conducted a keyword-based approach to create the Pars-HaO dataset, consisting of 8,013 tweets. Their result demonstrated that the BERT with the BiLSTM technique yielded the best outcome, achieving a macro F1-score of 70%.

While valuable, this line of inquiry suffers from some weaknesses. First, it is a small body of research, and we need more studies to enhance our understanding of the automated detection of incivility in Persian data. Second, as apparent in the papers reviewed above, these results are sometimes inconsistent. It makes the need for more research on this topic even more essential. In addition, most of these studies use lexicon-based or keyword-based approaches to collect data (Safayani et al., 2024; Sheykhlan et al., 2023). While these approaches could be beneficial in collecting hateful messages with specific words, they overlook uncivil tweets that might employ more implicit, i.e., rhetorical forms (Stoll et al., 2020). It could jeopardize the reliability of these datasets and methods, as they could not include all types of offensive messages. On a relevant point, the existing literature, as we discussed above, mainly distinguishes between offensive vs. non-offensive messages. This is also true for some works on the Persian language. Such an approach escalates the fallacy of overlooking implicit HS. Implicitly is a critical aspect of incivility (Stoll et al., 2020) , and we will exclusively analyze it in this research. Moreover, these studies compare DL, ML, and other computational methods without comparing their results with advanced LLM models. The only studies that embed ChatGPT in their experiments (Anonymous, 2025; Delbari et al., 2024) only used early versions of ChatGPT and did not analyze later and more sophisticated models like o1 or 4.1.

Finally, computational methods, particularly supervised models like BERT, need training data to be used in annotating unseen data. The quality of the training dataset is critical for the reliability of BERT output (Stoll et al., 2020). Nevertheless, creating such training data often involves extensive manual coding procedures (Witten, Frank, Hall, & Pal, 2016). Consequently, creating high-quality training data quickly becomes an expensive endeavor, which is probably why some studies rely on small sample sizes (Su et al., 2018). Further, Ross et al. (2017) have shown that the manual coding of hate speech requires clear definitions and guidelines to produce reliable annotations. Since labeling large datasets is expensive, many studies have used non-experts, such as crowd workers, to label their data (Davidson et al., 2017; Hsueh, Melville, & Sindhwani, 2009). In addition, Albladi et al. (2025) show that descriptions of the datasets used in LLMs are frequently absent or incomplete for hate speech detection through a comprehensive review. Considering works on Persian data, only a few public and semi-large datasets are available (Delbari et al., 2024). In addition, works on Persian data mainly employ English prompts (Anonymous, 2025; Delbari et al., 2024), another constraint limiting their findings.

We try to address these fallacies by providing a large dataset of Persian tweets and comparing human-driven qualitative interpretations with different SML models and LLMs. As will be discussed in the next section, we will follow a rigorous discursive and qualitative analytical approach to provide a rich and multi-class dataset of hateful messages. We will use ParsBERT from ML models, as previous research shows it worked well on Persian data. Since there is a dearth of research on more advanced LLMs, particularly ChatGPT models, in HS detection, we extend our analyses to different ChatGPT models. We specifically will use gpt-4.1, gpt-4.5-preview, gpt-4o and o1, gpt-4o-mini, o1-mini and o3-mini. We will also test

English and Persian versions of the same prompt for all models. Therefore, the following questions guide this research:

RQ1: How does ParsBERT perform in understanding different types of Persian uncivil messages compared to the human-driven qualitative method? Could this model be improved by strategies like weak labeling?

RQ2: How do different ChatGPTs (as listed above) perform in understanding different types of Persian uncivil messages compared to the human-driven qualitative method and ParsBert?

RQ3: To what extent could ParsBERT and ChatGPTs identify implicit Persian hateful messages?

RQ4: Does the language of the prompt affect the output of ChatGPTs, and if so, how?

**Context, data, and methods**

*Data collection*

In order to create our dataset, we focus on the #MahsaAmini movement., ika., Women, Life, Freedom, on Persian Twitter. WLF was a large-scale protest against the Islamic regime in Iran, sparked by the murder of Mahsa Amini for 'violating hijab' by the so-called 'morality police (Fadaee, 2024). While we only focused on this movement to collect the tweets dataset, more information to better understand the context in which this nationwide protest happened in Iran is presented in Appendix 4.a. In times of crisis, more hateful messages could be communicated on social media (Chekol et al., 2023). Therefore, we collected our data at a time that is plausible for a larger volume of uncivil tweets, unlike previous works, which randomly collected some data (Delbari et al., 2024; Sheykhlan et al., 2023). We have collected all popular Persian tweets, i.e., tweets with 1k+ likes/day, from September 15 to November 15, 2022, during the first two months of the movement using Twitter Academic API. Our data collection resulted in 47,278 tweets. We avoided keyword-based approaches and collected all popular tweets to be analyzed in the following steps. Therefore, our data collection consists of all kinds of tweets, hateful or non-hateful, and our dataset is potentially more diverse than those in previous research on Persian data (Mozafari et al., 2024; Safayani et al., 2024).

*Human-driven qualitative analysis*

We followed a rigorous qualitative and discursive approach to code the dataset. We drew on the Social Media Critical Discourse Studies (SM-CDS) approach (KhosraviNik, 2017). SM-CDS includes a horizontal context substantiation, which deals with the intertextuality among textual practices, and a vertical context substantiation, which links both the micro-features of textual analysis and the horizontal context to the sociopolitical context of users in society. SM-CDS enables us to discursively investigate users' and tweets' metadata simultaneously to reach the deeper layers of shaping uncivil messages on Twitter.

Five coders were trained to code tweets following our analytical approach. They read each tweet closely, consider the author and tweet metadata, e.g., images or videos, and refer to Twitter where necessary to gain more knowledge about the text under investigation. Coders coded two variables: Incivility type and the level of implicity (1 for the most implicit and 3 for the most explicit messages). When they coded 50% of the dataset, we calculated the intercoder agreement (F1-score), 0.74 and 0.79 for the two variables, respectively. Then, we discuss the discrepancies and disagreements in a meeting. Next, coders coded the whole dataset from the beginning. The final F1-score increased to 0.83 and 0.86, respectively. We had another meeting, decided on disagreements, and coded them based on the consensus agreement.

*ParsBERT*

As discussed earlier, we employed ParsBERT as a supervised model, which performed well on Persian data (Delbari et al., 2024). ParsBERT is a monolingual language model based on the BERT architecture (Farahani et al., 2021). It is pre-trained on more than 3.9 million Persian documents, allowing it to capture contextual representations of Persian text. We composed a training dataset from the main dataset, as will be discussed in the Findings section. We also augmented the training dataset using weak labeling to test the effectiveness of the improved training dataset. Weak labeling is a form of data annotation in which the labels used for training machine learning models are noisy, incomplete, or imbalanced (Zhou, 2018). This method automates the creation of labeled training data by aligning unstructured text with existing annotated data.

*ChatGPT models*

As we mentioned in this paper, we employed seven different ChatGPTs with both English and Persian prompts. Each model has its particular architecture and capabilities, as presented in Appendix 1. We compare the models based on their accuracy (F1-score), time, and the number of tokens used. In addition, we did not use the whole coded dataset in this step, given that it would require a lot of time and money. We constructed a dataset of 2,086 tweets, which we will explain in the Findings section, to be used in this phase. Models were tasked with identifying the incivility type and level of implicity (LoI). Each model was instructed to assign a LoI score on a three-point scale (1 = high and 3 = low), while non-hateful content was assigned a score of 0. We also asked the models to briefly explain the rationale behind their decisions. Both English and Persian prompts are presented in Appendix 2.

**Findings**

Table 1 presents the result of human-driven qualitative analysis, which shows the percentage of hateful and non-hateful messages in the dataset.

| Incivility type | Percentage |
| --- | --- |
| Pejorative speech | 15 |
| Insult | 13 |
| Threatening messages | 3,5 |
| Neutral (not uncivil)[2] | 68,5 |

Table 1. The percentage of hateful and non-hateful messages in the dataset

Table 1 reveals that 68,5 percent of tweets were not uncivil, and as a result, only 31,5% were hateful. As expected, the percentage of hateful tweets decreased as the severity of the uncivil messages rose. The following table indicates the percentage of LoI in uncivil tweets.

| Level of Implicity | Percentage |
| --- | --- |
| 0 (non-hateful) | 68.5 |
| 1 (high) | 1 |

---

[2] We used terms like Neutral, not-uncivil and non-hateful to refer to messages were not coded as hateful/uncivil.

| 2 (medium) | 11.7 |
| 3 (low) | 18.8 |

Table 2. The percentage of implicit hateful messages

Table 2 shows that the distribution of LoI was also highly imbalanced. The majority of uncivil tweets are shared between 2 and 3 classes. Thus, the percentage of highly implicit tweets was small. These findings also indicate that most users expressed their uncivil sentiments explicitly in direct ways. Thus, the problem of implicit uncivil messages may not be as critical as suggested by studies like Stoll et al. (2020) or Frenda et al. (2023).

To gauge the capability of ParsBERT in the detection of incivility, we created a sample of the whole dataset for the training dataset. As seen above, the majority of messages in the original dataset were non-uncivil, 68,5%. Thus, it made the dataset highly imbalanced, which cannot be used perfectly with supervised learning. As a result, we kept all uncivil messages but only included a random sample of 6000 non-hateful tweets in the training dataset to make it more balanced. The research sample A, which we also used for testing ChatGPTs, included: pejorative speech: 6923; insult: 6263; neutral: 6000; and TMs: 1675.

Table 3 reveals how the ParsBERT model performs in identifying uncivil tweets.

| Category | Precision | Recall | F1-Score | Support |
|---|---|---|---|---|
| Neutral | 0,79 | 0,8 | 0,8 | 1200 |
| Insult | 0,68 | 0,71 | 0,69 | 1253 |
| Pejorative speech | 0,67 | 0,65 | 0,66 | 1385 |
| Threatening messages | 0,43 | 0,37 | 0,4 | 335 |
| Accuracy | | | 0,69 | 4173 |
| Macro avg | 0,64 | 0,63 | 0,64 | 4173 |
| Weighted avg | 0,69 | 0,69 | 0,69 | 4173 |

Table 3. The result of ParsBert in identifying HS in research sample A

Table 3 indicates that the F1-score weighted was 0,69, which is acceptable but not strong enough. The model worked well in identifying neutral, insult, and PS classes, but notably worked weakly in the identification of TMs (F1-score of 0,4). This result could be pertinent to the small number of TMs in the training dataset. Using weak labeling, we augmented the training dataset by running the fine-tuned model on a sample of 100,000 unseen data. We extracted this data from tweets shared by Iranian users during the same time as the research period. The model returned 4,738 TMs, which were checked by human coders, and then 4,085 of them were added to the training dataset. As a result, the augmented dataset had 5,760 TMs, making the training dataset more balanced. Then, we conducted ParsBERT again using the new training dataset (research sample B). The output is presented in Table 4.

| Category | Precision | Recall | F1-Score | Support |
|---|---|---|---|---|
| Neutral | 0,82 | 0,8 | 0,81 | 1200 |
| Insult | 0,8 | 0,8 | 0,8 | 1253 |
| Pejorative speech | 0,8 | 0,8 | 0,8 | 1385 |
| Threatening messages | 0,72 | 0,7 | 0,71 | 1152 |
| Accuracy | | | 0,8 | 4990 |
| Macro avg | 0,79 | 0,78 | 0,78 | 4990 |
| Weighted avg | 0,79 | 0,78 | 0,78 | 4990 |

Table 4. The result of ParsBert in identifying HS in research sample B

Table 4 reveals that the F1-score significantly increased to 0.78. This finding shows that using a balanced training dataset plays an important role in using BERT.

Now, we present the results on the strength of ChatGPTs in the automated detection of incivility to answer RQ2. Using quota sampling, we constructed a new dataset by selecting 10 % of the research sample A. The resulting subset contains 2,086 instances and preserves the original class distribution of sample A.

First, we present the tokens and the time each model used in Table 5.

| Model | Prompt | Total tokens used | Time spent (minutes) |
|---|---|---|---|
| 4o | Farsi | 2.703.270 | 55 |
| 4o | English | 1.936.230 | 67 |
| o4-mini | Farsi | 2.676.995 | 64 |
| o4-mini | English | 1.976.000 | 58 |
| 4.5-preview | Farsi | 2.701.460 | 255 |
| 4.5-preview | English | 1.918.763 | 262 |
| o1 | Farsi | 6.598.221 | 1011 |
| o1 | English | 6.933.813 | 3549 |
| o1-mini | Farsi | 4.270.240 | 223 |
| o1-mini | English | 4.224.480 | 251 |
| o3-mini | Farsi | 5.663.840 | 531 |
| o3-mini | English | 4.374.240 | 477 |
| 4.1 | Farsi | 3.107.160 | 117 |
| 4.1 | English | 2.215.921 | 61 |

Table 5. The number of tokens and time used by ChatGPTs

Table 6 reveals that the models usually consumed more tokens when they used the Farsi prompt. The only exception is o1, which lightly consumed fewer tokens with the Farsi prompt. In addition, it took less time for most of the models to annotate the dataset with the Farsi prompt. Of course, there are some exceptions here, too: 4o-mini, o3-mini, and gpt-4.1. In terms of time, o1 acted remarkably faster with Farsi than with English prompts. Also, this model used the highest number of tokens of the tested models. Since o1 is a reasoning model with full capacity, it was expected. o1-mini and o3-mini also used more tokens than the other models. However, OpenAI pricing is different for these models and is much cheaper for these two models than 4o or 4.5-preview. The o1 and 4.5-preview models are the most expensive models. From these models, o1 also consumed a considerable number of tokens and took a long time. Regarding the time consumed, 4o, o4-mini, and 4.1 models are the fastest. It should also be mentioned that the 4.1 model needs less money in comparison with most of the large models, which makes it a promising, fast, and less expensive model (see Appendix 1). However, these factors should be considered in line with each model's accuracy. Table 6 provides us with the level of accuracy of these models in HS identification. After analyzing this variable, we will be able to better evaluate each model's power and shortcomings.

| Model Name | Prompt Language | F1-Score (weighted) |
|---|---|---|
| o4-mini | English | 0,49 |
| o4-mini | Farsi | 0,47 |
| o3-mini | English | 0,49 |
| o3-mini | Farsi | 0,44 |
| o1-mini | English | 0,49 |
| o1-mini | Farsi | 0,49 |
| o1 | English | 0,53 |
| o1 | Farsi | 0,52 |
| 4o | English | 0,5 |
| 4o | Farsi | 0,51 |
| 4.5 preview | English | 0,56 |
| 4.5 preview | Farsi | 0,56 |
| 4.1 | English | 0.57 |
| 4.1 | Farsi | 0.56 |

Table 6. F1-score (weighted) for ChatGPTs identifying the type of incivility

Table 6 reveals that mini models (o1, o3, and o4) work relatively weaker than the other big models. In addition, there are no remarkable differences between F1-scores for English and Farsi prompts. The models returned almost the same results for both prompts. This table also shows that the gpt-4.5-preview and 4.1 models performed better in HS detection. The F1-scores for these models are almost 0.56, but still lower than the BERT model. Thus, it is plausible to argue that ChatGPT models still fall behind BERT.

We created a confusion matrix for each model to understand the codes where models caused the most errors in data annotation. The confusion matrices are presented in Appendix 3 (Figures 1-14). We investigate the confusion matrices to find where the models' codes differ vastly from those of humans. In this step, we only focus on the gpt-4.5-preview, gpt-4.1, and gpt-o1 models due to their higher performance. Table 7 below presents the results of this step.

| Model | Prompt | Problematic codes (Human- model) |
|---|---|---|
| 4.5 Preview | Farsi | PS - Neutral / Insult- PS |
| 4.5 Preview | English | PS - Neutral / Insult – PS |
| o1 | Farsi | PS - Neutral / Insult |
| o1 | English | PS - Neutral / Insult |
| 4.1 | English | Neutral- PS / Insult- PS |
| 4.1 | Farsi | Neutral- PS / Insult- PS |

Table 7. The most problematic codes in HS detection by ChatGPT models

Table 7 indicates that each model's behavior was the same for both Farsi and English prompts. For instance, the o1 model returned most errors for tweets that were coded as PS by humans and Neutral/Insult by the model for both prompts. While models were mainly challenged by PS, the TMs were not difficult to identify. We analyzed tweets where these errors happened and read ChatGPT's reasons to understand why they happened. As apparent in the above table, the main problem for all models was coding PS, which they labeled as Neutral or vice versa (in the 4.1 case). Our discursive analyses show that the models still struggle with the identification of incivility when it is more indirect and discursive. For instance, GPT-4.5-preview (with both Farsi and English prompts) annotated the following tweet as neutral: *"Let the world remember — we are fighting a regime that has captured a group of elite students in a university parking lot and wants to slaughter them! This is the true and clear image of the Islamic Republic!"* This tweet clearly depicts the Iranian regime as a brutal and violent system, which is going to *slaughter* elite students. Thus, it is hateful and shows clear disparagement against the regime, which was correctly coded by humans as PS. In another tweet, the author employed *#suppression_republic* to pejoratively refer to the Iranian regime. However, this model again coded it as Neutral. The model responses also were not helpful, as they simply argue there is no sign of humiliation or disparagement in tweets. More examples and explanations are presented in Appendix 4.b.

The other reason for the error in the models' codes is the fact that human coders considered all metadata and replies for coding tweets, while the models only accessed the tweets' text. For instance, the same model coded this tweet as Neutral, which makes sense: "It was beautiful. I liked it." The text itself contains no hateful or uncivil words. However, it is attached to an image that includes a sentence by the Iranian leader to humiliate and denounce protesters and others who want to overthrow the regime (see the image below). Therefore, the tweet in fact is pejorative, but the model is not reasonably able to understand it.

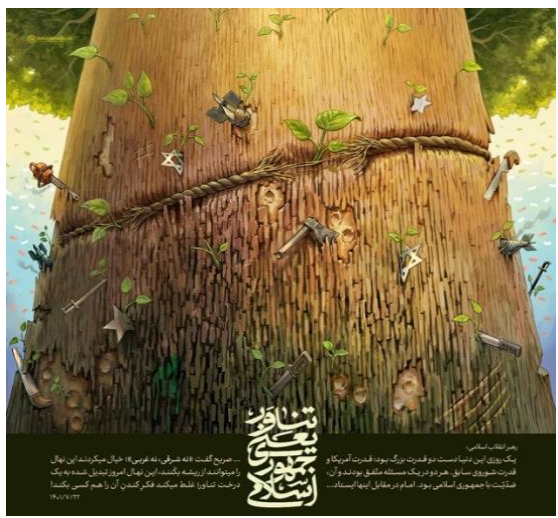

Image 1. The pejorative image embedded in a seemingly non-uncivil tweet

This is also true for tweets, which are reports from the streets, including videos and images. In such cases, the video/image includes hateful content, but the text itself is more in the form of an objective report. While in such cases it is understandable why the model made errors, there are cases where clear hateful or disparaging words are embedded in the text, but the models, again, failed to identify the correct type. For instance, a tweet was written about *Islamic Republic forces fleeing* and shared a video. This text has a pejorative tone against the Iranian forces; however, the models failed to detect it. Nonetheless, the frequency of these errors is lower than that of errors that occurred due to the inability of models to understand more complicated discursive layers of text. We should also add that if we had only focused on coding tweets based on the text and removing other tweets including pictures/videos, the problematic messages could have been lower. However, we wanted to test whether the models are able to detect the true label only based on the text and the contextual data embedded in large language models. In order to minimize the effect of such messages, we removed them from the dataset and then ran gpt-4.1(English) as the best model. The result shows that the effect of these messages was not high, and the F1-score only rose by 0.02, so the final score increased to 0.59.

In the next step, we examined models' ability to identify implicit uncivil messages. We should add that we were not able to run ParsBERT in this step because the code classes were highly imbalanced, as indicated in Table 2. The next table presents the results of our tests with ChatGPTs.

| Model | Prompt | F1-score (weighted) |
|---|---|---|
| o4-mini | English | 0.55 |
| o4-mini | Farsi | 0.54 |
| o3-mini | English | 0.52 |
| o3-mini | Farsi | 0.58 |
| o1-mini | English | 0.57 |
| o1-mini | Farsi | 0.57 |

| o1 | English | 0.60 |
| o1 | Farsi | 0.61 |
| 4o | English | 0.55 |
| 4o | Farsi | 0.57 |
| 4.5 preview | English | 0.59 |
| 4.5 preview | Farsi | 0.58 |
| 4.1 | English | 0.62 |
| 4.1 | Farsi | 0.60 |

Table 8. F1-score (weighted) for ChatGPT models identifying LoI

Table 8 reveals that ChatGPTs performed relatively better in identifying implicity than the type of HS. Again, the performance of mini models was generally weaker than that of other models in the detection of implicity. However, it should be added that mini models also worked remarkably better in this task. The only exception among big models is o4, whose performance is as weak as that of mini models. It also resembles the weak performance of o4 in identifying HS type (Table 6).

In addition, most of the models performed almost identically with Farsi and English prompts. For instance, the F1-score for models like o4-mini, o1-mini, and 4.5 preview was the same for both prompts. In other cases where the scores are not completely the same, there is no pattern. Thus, we cannot conclude if the models returned more reliable results with any of these languages. For instance, there is a discrepancy in the o1 and 4.1 models' results, which returned the highest scores, on working with different languages. o1 worked better with Farsi, while 4.1 performed better with the English prompt.

Like we did for the type of incivility, we created and investigated confusion matrices for the models (gpt-o1 and gpt-4.1, due to their high performance) to understand where they returned the most errors. The confusion matrices are presented in Appendix 3 (Figures 15-28). Table 9 presents the results of our analyses.

| Model | Prompt | Problematic codes (Human-model) |
|---|---|---|
| o1 | Farsi | 3-2 |

| | | 2-0 |
|---|---|---|
| o1 | English | 3-2 |
| | | 2-0 |
| 4.1 | Farsi | 3-2 |
| | | 2-0 |
| 4.1 | English | 2-3 |
| | | 2-0 |

Table 9. The most problematic codes in detection of implicit uncivil messages by ChatGPT models

Table 9 shows that one of the most problematic codes is 3-2, which is not a surprise since the agreement about the severity and implicity of uncivil messages could be challenging while the categories are close. For instance, gpt-4.1 model (Farsi) coded the implicity of the below tweet with 2.

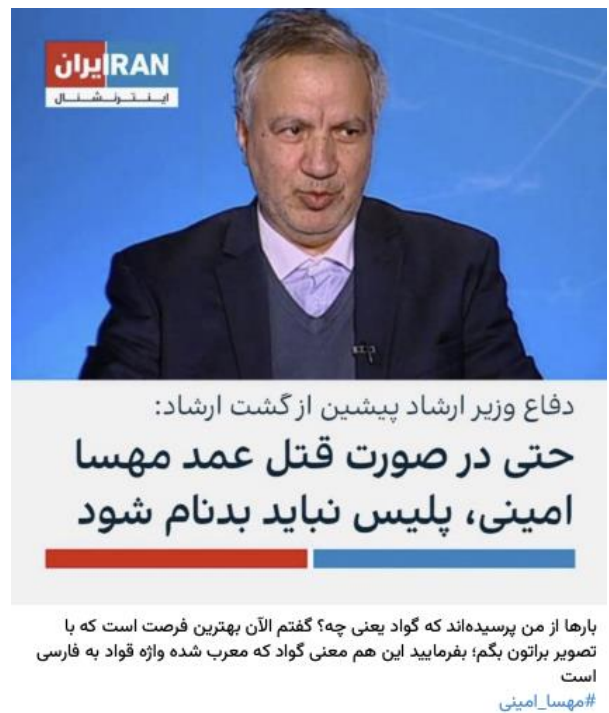

Image 2. Tweet example of disparity in coding implicity

The picture in this tweet includes a quote from the former Minister of Culture in Iran, Ata'ollah Mohajerani: *Even if Mahsa Amini was deliberately killed, the police should not be disgraced*. Then the tweet author wrote: '*I've been asked many times what "goad" means. I thought now is the best time to show you with a picture. Here you go — this is the meaning of goad, which is the Arabicized form of the Persian word qavvād (قواد).*' *Qavvād* is used as a synonym of pimp in Farsi and arabic. In this tweet, this metaphor was used to humiliate and insult Mohajerani in a clear way. Thus, the LoI is coded correctly by human coders with 3.

However, the error that happened in 2-0 was unexpected. Such errors are a result of what we have discussed before. The model failed to understand that some messages are uncivil even when there are some clear bad words. For instance, an anti-regime account posted a video and wrote: *A visual account of the savagery of the Islamic Republic's mercenaries*, which clearly is an explicit, uncivil message. Surprisingly, the 4o model did not detect any hateful content and explained: *This tweet contains no explicit or uncivil content, as it merely refers to a depiction of the current situation*. It is not clear how the model did not consider savagery and mercenaries as uncivil words. The instances of 2-0 error indicate that the models' main problem in identifying implicit uncivil messages is still rooted in their inability to identify HS. Even with explicit and clear bad words, it happened to models to misclassify tweets, and as a result, misclassify the LoI. However, the 4.1 model worked better in this task since the frequency of this error is not high. This model mainly made mistakes between close categories, not between clear and ambiguous uncivil tweets. More examples of and explanations about such errors are presented in Appendix 4.b.

**Discussion**

This paper enhances our understanding of automated HS detection, particularly on an understudied language: Farsi. Given the focus of the main strand of research on languages like English (MacAvaney et al., 2019), this study tests different methods to evaluate their capability in identifying incivility on Persian tweets. Empirically, we provide the largest public dataset of uncivil Persian tweets. The new dataset could be used by other researchers in monolingual or multilingual text analysis to advance this line of inquiry further.

Moreover, we examined LLMs, ChatGPTs in this case, BERT, and human-driven qualitative coding to identify different classes of incivility to push forward mixed methods research in this field, which focuses on binary classification of hateful/non-hateful categories to a large extent (Pendzel et al., 2023; Struppek et al., 2024). In addition, we treat implicity as an independent variable, recognizing that its degree may vary across different categories of incivility. Last but not least, we tested prompts with both English and Farsi to test their effect on the ChatGPTs' output.

This study's main finding is that SML models, particularly ParsBERT, performed significantly better than ChatGPTs in HS detection. ParsBERT F1-score with research sample A, which was imbalanced, was 0,69, which rose to 0.78 after balancing the sample. The best ChatGPT model, gpt-4.1(English), returned an F1-score of 0.57. These findings echoed previous research, which employed early versions of ChatGPT like GPT-3.5-turbo (Delbari et al., (2024). Thus, the advancement of models did not notably affect their reliability in HS identification on Persian data. However, some studies report high F1-scores for this task (Deroy & Maity, 2023; Leidinger et al., 2023). The contradiction here may be the result of the fact that such studies mainly focus on a binary classification of hateful messages. Therefore, when we consider different types of incivility, the performance of ChatGPT may not be as high as its work on binary classes. Given previous research on Persian data in addition to our findings, we can conclude that LLMs and ChatGPT in particular are not the best choices for HS detection despite the boom around them. Using SML models like BERT, and its variations (Delbari et al., 2024; Safayani et al., 2024), or some ensemble models like what Mozaffari et al. (2024) proposed, could return more reliable output.

Findings also reveal that the performance of ChatGPTs in the detection of implicity is better than HS. Models like o1 and 4.1 returned F1-scores of about 0.60. However, discursive interpretations of the most problematic codes show that the main problem here is also the models' inability to detect HS. In fact, models work roughly acceptably on detecting implicity, but errors arise from the fact that they cannot precisely distinguish between non-hateful messages and different types of incivility (mainly PS). ChatGPTs even failed to identify the correct type of incivility when the tweet includes some clear bad words. While the fact that these models work less well on identifying more discursive or rhetorical forms of incivility is not surprising, their inability to detect clear uncivil messages is unexpected. That is true that we fed the models with the text of tweets, even where tweets include visual elements like video. However, the effect of such messages was minor. The primary limitation of the models may stem from the contextual data used during their training. However, we cannot be sure about that due to our lack of access to this data.

From ChatGPTs, mini models mainly work worse than other big models, particularly in HS detection. At the same time, some of them, like o1-mini and o3-mini, consumed a large number of tokens. o1 model, despite its relatively good performance, used the most tokens and took the longest time to code data. Given its high price and the fact that the 4.1 and 4.5 preview models' performance was the same or better, o1 is not a reasonable choice for coding data. The behaviour and performance of the best two models, 4.1 and 4.5 preview, were highly identical. 4.1 is a newer model (till April 2025), which our test shows returned the same result in less time. It also needs significantly less money than 4.5-preview, which is a very expensive model. However, while we can say that the time and money consumed by newer models, i.e., 4.1, reduces, their reliability does not increase significantly. Finally, our findings reveal that the prompt's language does not affect the output notably.

While this paper enhances our understanding of automated detection of incivility, there is still some room for improvement. The most significant limitation of our research is that we could not run ParsBERT to detect LoI. Further studies could provide more balanced training datasets to include this model in the tussle as well. In addition, our investigation is limited to Persian as an underexplored language. Scholars could draw on this inquiry to conduct multilingual analyses or test LLMs on different languages, including Farsi. Moreover, we fed SML and ChatPPTs only with textual data, while human coders also considered the metadata, such as pictures and videos, in their coding. Despite our analyses, which show the effect of such messages was minimal on ChatGPT's reliability, further research needs to be conducted on multimodal analyses considering text and other data to minimize such an effect. Last but not least, LLMs are developing rapidly, and new models are introduced from time to time. Researchers should consider new models and compare them to understand if the developments lead to an acceptable performance in HS detection. Based on the thorough analysis we provide in this research, we suggest using the latest ChatGPT models and comparing them with other LLMs. Therefore, no need to run analyses with older or mini models again.

**Data availability statement:** The authors confirm that the data supporting this study's findings are available within the supplementary materials. The data will be made publicly available upon publication of the research.

Guo, K., Hu, A., Mu, J., Shi, Z., Zhao, Z., Vishwamitra, N., & Hu, H. (2024). *An Investigation of Large Language Models for Real-World Hate Speech Detection* (Version 1). arXiv. https://doi.org/10.48550/ARXIV.2401.03346

Hameleers, M., Van Der Meer, T., & Vliegenthart, R. (2022). Civilized truths, hateful lies? Incivility and hate speech in false information – evidence from fact-checked statements in the US. *Information, Communication & Society*, *25*(11), 1596–1613. https://doi.org/10.1080/1369118X.2021.1874038

Hietanen, M., & Eddebo, J. (2023). Towards a Definition of Hate Speech—With a Focus on Online Contexts. *Journal of Communication Inquiry*, *47*(4), 440–458. https://doi.org/10.1177/01968599221124309

Huang, F., Kwak, H., & An, J. (2023). Is ChatGPT better than Human Annotators? Potential and Limitations of ChatGPT in Explaining Implicit Hate Speech. *Companion Proceedings of the ACM Web Conference 2023*, 294–297. https://doi.org/10.1145/3543873.3587368

Keipi, T., Näsi, M., Oksanen, A., & Räsänen, P. (2016). *Online Hate and Harmful Content: Cross-national perspectives* (1st ed.). Routledge. https://doi.org/10.4324/9781315628370

Kovács, G., Alonso, P., & Saini, R. (2021). Challenges of Hate Speech Detection in Social Media: Data Scarcity, and Leveraging External Resources. *SN Computer Science*, *2*(2), 95. https://doi.org/10.1007/s42979-021-00457-3

Leidinger, A., van Rooij, R., & Shutova, E. (2023). The language of prompting: What linguistic properties make a prompt successful? In H. Bouamor, J. Pino, & K. Bali (Eds.), *Findings of the Association for Computational Linguistics: EMNLP*